\documentclass{article}
\title{IIP-Mixer: Intra-Inter Patch Mixing Architecture for Battery Remaining Useful Life Prediction}
\author{Guangzai\ Ye$^{1}$, Li\ Feng$^{1}$\thanks{Corresponding author:lfeng@must.edu.mo}, Jianlan\ Guo$^{2}$, Yuqiang\ Chen$^{2}$\\
$^{1}$School of Computer Science and Engineering,\\Macau University of Science and Technology, Macau SAR, China\\
$^{2}$Dongguan Polytechnic, Dongguan, China}
\date{}

\usepackage{csquotes}

\usepackage{multirow}
\usepackage{multicol}

\usepackage{booktabs}

\usepackage{amsmath}
\usepackage{amssymb}

\usepackage{graphicx}
\graphicspath{{images/}}

\usepackage[sorting=none]{biblatex}
\begin{document}
\maketitle
\begin{abstract}
 Accurately estimating the Remaining Useful Life (RUL) of lithium-ion batteries is crucial for maintaining the safe and stable operation of rechargeable battery management systems. However, this task is often challenging due to the complex temporal dynamics. Recently, attention-based networks, such as Transformers and Informer, have been the popular architecture in time series forecasting. Despite their effectiveness, these models with abundant parameters necessitate substantial training time to unravel temporal patterns. To tackle these challenges, we propose a simple MLP-Mixer-based architecture named \enquote{Intra-Inter Patch Mixer} (IIP-Mixer), which is an architecture based exclusively on multi-layer perceptrons (MLPs), extracting information by mixing operations along both intra-patch and inter-patch dimensions for battery RUL prediction. The proposed IIP-Mixer comprises parallel dual-head mixer layers: the intra-patch mixing MLP, capturing local temporal patterns in the short-term period, and the inter-patch mixing MLP, capturing global temporal patterns in the long-term period. Notably, to address the varying importance of features in RUL prediction, we introduce a weighted loss function in the MLP-Mixer-based architecture, marking the first time such an approach has been employed. Our experiments demonstrate that IIP-Mixer achieves competitive performance in battery RUL prediction, outperforming other popular time-series frameworks.
 \end{abstract}
 \section{Introduction}
 Lithium-ion batteries are widely used in electric vehicles, unmanned aerial vehicles, and grid energy storage systems. Accurately predicting the Remaining Useful Life (RUL) of a battery plays an important role in managing the health and estimating the state of a lithium-ion battery. \cite{alsuwian2024review} \cite{wei2021remaining}
However, accurately predicting the RUL of batteries is highly challenging due to nonlinear degradation mechanisms caused by cycling and varied operational conditions. Generally, battery RUL is defined as the number of cycles when a battery reaches 80\% initial capacity. Previous works on battery RUL prediction can be categorized into two main approaches: physics-based approaches and data-driven approaches. \cite{xu2023hybrid}

Physics-based approaches such as the single particle model \cite{santhanagopalan2006review} and pseudo-two-dimensional model \cite{kemper2015simplification} based on electrochemical principles underlying Lithium-Ion Batteries (LIBs) can simulate a battery’s current and voltage characteristics from kinetics and transport equations. Although these models are usually precise and understandable, they come with significant computational demands and difficulties when extending their applicability to different battery cell types. \cite{wang2022data}

Conversely, data-driven approaches do not make a priori assumptions about battery degradation mechanisms but instead leverage historical cycling data of batteries. Most of the current data-driven-based approaches integrate intricate structures such as gating mechanisms \cite{wang2022data} \cite{wang2023improved} and attention mechanisms \cite{zhang2023data} \cite{chen2022transformer}. Though effective, these complex approaches require significant training time to thoroughly explore temporal and intercorrelation patterns, demanding extensive computing resources. \cite{ye2023prediction} Besides, the permutation-invariant self-attention mechanism leads to a loss of temporal information to some extent. \cite{zeng2023transformers} Our work aims to design a simple model that does not rely on any form of costly complex gates operation, or attention mechanism, yet performs competitively compared to existing models. 

In this work, we introduce the Intra-Inter Patch Mixer (or \enquote{IIP-Mixer} for short) architecture. This architecture is exclusively based on the application of multi-layer perceptron across intra-patch and inter-patch. Importantly, IIP-Mixer refrains from employing gating, convolutions \cite{wei2024state}, or self-attention mechanisms, relying solely on basic matrix multiplication routines, changes to data layout (reshapes and transpositions), and scalar nonlinearities. \cite{tolstikhin2021mlp} Our contributions are the following:
\begin{itemize}
\item We propose IIP-Mixer, a novel model built on all MLP architecture that leverages intra-inter patch mixing MLP design to capture the local and global temporal patterns in the time series.
\item To achieve precise predictions of Remaining Useful Life, we introduce the parallel dual heads to aggregate output information from both intra-patch mixing MLP and inter-patch mixing MLP. 
\item Considering the varying importance of different variables in battery Remaining Useful Life prediction, we proposed a weighted loss function to further enhance prediction performance.
\item The experimental results show that the IIP-Mixer architecture proposed can effectively improve the accuracy of the prediction of battery remaining useful life relying solely on MLP-based structures.
\end{itemize}

 \section{Related Work}
As rechargeable batteries are a source of power for many devices, such as electric cars, mobile phones, unmanned aerial vehicles, and so on, it is critical to ensure their reliability and safety. Remaining useful life predictions have become increasingly important topics and have received considerable attention in recent years. To predict RUL for rechargeable batteries accurately, many deep-learning models have been proposed. \cite{liang2024hybrid}
\begin{itemize}
\item MLP: Multi-layer perceptron (MLP) is a kind of artificial neural network with a forward structure, including an input layer, an output layer, and several hidden layers \cite{wu2016online}. For RUL prediction, MLP is applied to learn the nonlinear capacity degradation of the battery. However, as the number of layers increases, the number of parameters increases rapidly, which easily leads to overfitting.
\item RNN: Recurrent neural network (RNN) is a kind of neural network for processing sequential data \cite{sutskever2014sequence} \cite{zhang2018long} \cite{shi2021dual}. Sharing parameters across different parts of the model makes it possible to apply the model to examples of different lengths. However, recurrent neural networks with a recurrent manner have a high time cost for training, and the performance will degrade due to long-term dependency.
\item Transformer: To learn trends from sequential data, the multi-head attention network of the Transformer is used to capture sequential information precisely and improves the training performance of neural networks \cite{chen2022transformer} \cite{vaswani2017attention}. However, for a small dataset, it can easily lead to overfitting. Therefore, the performances of Transformer-based models are not always greater than MLP-based or RNN-based frameworks. 
\end{itemize}

\section{Methods}
\subsection{Problem Formulation}
For systems like lithium-ion batteries, the degradation process often spans lots of cycles, with numerous variables collected by various sensors during each cycle. In this work, the battery RUL prediction problem is a case of multivariate time series prediction problem. Specifically, given the historical observations $\boldsymbol{X}\in\mathbb{R}^{C\times L}$, where $L$ is the length of the lookback window, $C$ is the number of variables. We consider the task of predicting $\boldsymbol{Y}\in\mathbb{R}^{C\times N}$, where $N$ is the number of subsequent time steps. \cite{chen2023tsmixer} In this work, we focus on the case when the values of the target time series are equal to the historical observation.

\subsection{The Framework}
Inspired by the recent MLP architecture MLP-Mixer \cite{tolstikhin2021mlp}, The idea behind the IIP-Mixer architecture is to separate the intra-patch mixing and inter-patch mixing to capture local temporal patterns and global temporal patterns simultaneously. Both operations are implemented with MLPs. The central architecture of the IIP-Mixer is an MLP-based design, which aggregates information from both intra-patch and inter-patch within the input series. It is worth noting that, different future has different patch mixers. For univariate time series, as illustrated in Figure 1, this framework contains the following major components: 
\begin{figure}[h]
    \centering
    \includegraphics[width=0.8\textwidth]{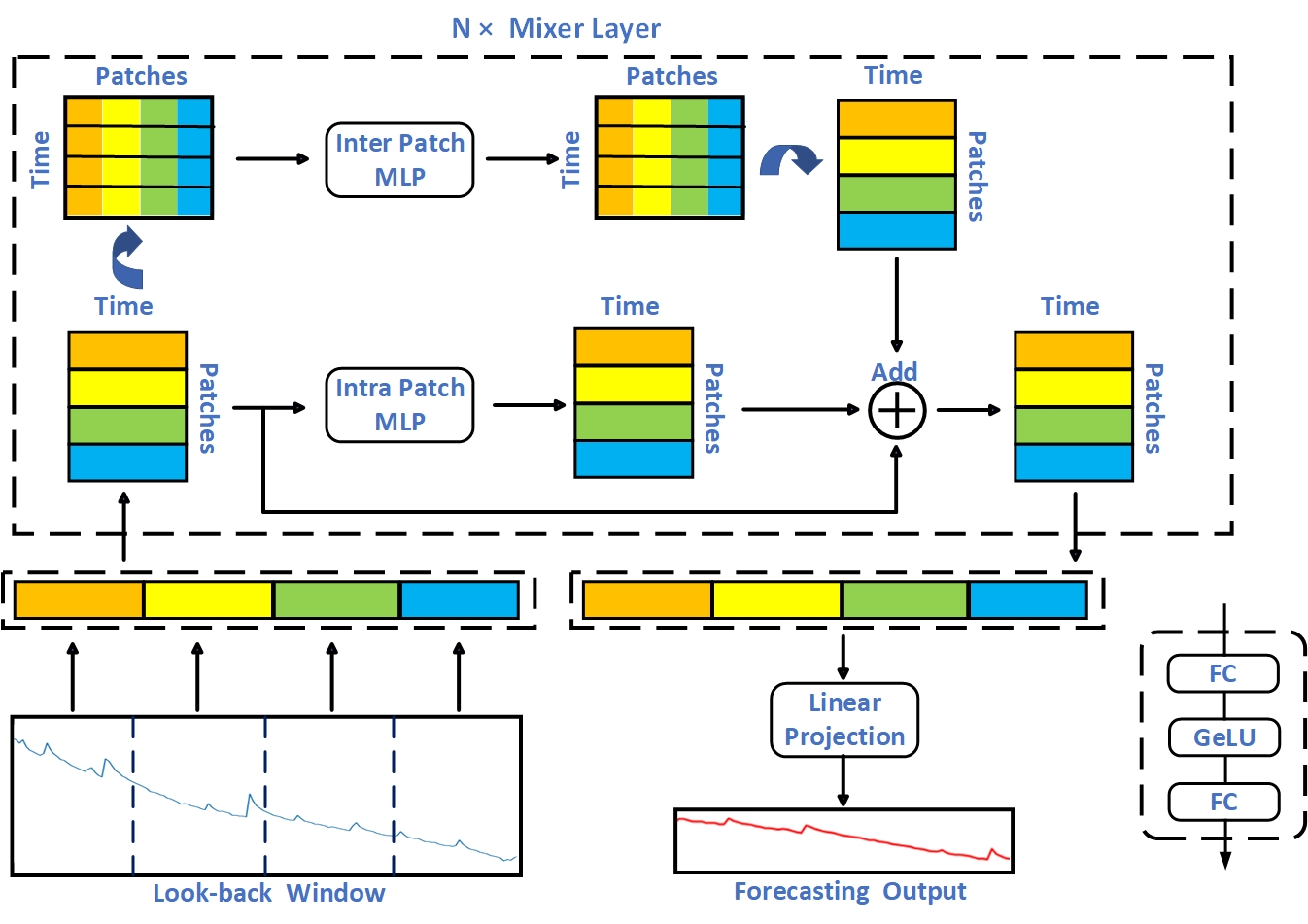}
    \caption{Intra-Inter Patch Mixing Architecture for Battery Remaining Useful Life Prediction.}
    \label{fig:IIMixing1}
\end{figure}

\textbf{Input series transformation:} From the perspective of channel independence, the multivariate time series $\boldsymbol{X} \in \mathbb{R}^{C \times L}$ were divided into $C$ univariate series $\boldsymbol{X}^{(i)} \in \mathbb{R}^L, \quad i=1,2, \cdots C$. These univariate series are independently fed into the IIP-Mixer model. This approach breaks down these input univariate series into smaller and structured patches. It transforms each original univariate time series from a 1D series to 2D patches while preserving their original relative positions, expressed as follows:
$$
\boldsymbol{X}^{(i)} \in \mathbb{R}^L \rightarrow \boldsymbol{X}^{(i)} \in \mathbb{R}^{H \times W}
$$
where $W$ represents the patch length, and $H$ denotes the number of patches for each univariate series, which is a tuning hyperparameter.

\textbf{Intra-patch mixing MLP:} The rows in the 2D input univariate series $\boldsymbol{X}^{(i)}$ represent distinct patches, while the columns denote time steps. A trainable intra-patch mixing MLP shared across all patches is employed to map each patch to a hidden space. We utilize a multilayer perceptron with a single hidden layer to capture local temporal patterns within patches. The size of output $\boldsymbol{O}_{\text {intra }}^{(i)}$ is the same with input $\boldsymbol{X}^{(i)}$, the process can be summarized as the following equations:
$$
\boldsymbol{O}_{\text {intra }}^{(i)}=\boldsymbol{W}_2 \sigma\left(\boldsymbol{W}_1 \boldsymbol{X}^{(i)}\right)
$$
It is essential to highlight that, the parameters of intra-patch mixing MLP shared across all patches prevent the architecture from growing too fast when increasing the length $L$ of the lookback window and leads to significant memory savings.

\textbf{Inter-patch mixing MLP:} The 2D patches $\boldsymbol{X}^{(i)^T}$ is a transposition of $\boldsymbol{X}^{(i)}$. Scilicet, the rows of which represent time steps, while the columns denote distinct patches. A trainable inter-patch mixing MLP shared across all of the time steps is employed to map each time step to a hidden space. The same as intra-patch mixing MLP, inter-patch mixing MLP utilizes a multilayer perceptron with a single hidden layer to capture global temporal patterns across patches. The size of output $\boldsymbol{O}_{\text {inter }}^{(i)}$ is the same with input $\boldsymbol{X}^{(i)^T}$, the process can be summarized as the following equations:
$$
\boldsymbol{O}_{\text {inter }}^{(i)}=\boldsymbol{W}_4 \sigma\left(\boldsymbol{W}_3 \boldsymbol{X}^{(i)^T}\right)
$$

\textbf{Linear projection:} Linear projection is a trainable linear neural network that projects aggregated output information from the intra-patch mixing MLP and inter-patch mixing MLP to predict future time steps. Additionally, the input of linear projection incorporates skip-connections of $\boldsymbol{X}^{(i)}$, This residual design ensures that IIP-Mixer retains the capacity of temporal linear models while still being able to exploit intra-inter patch information. The linear projection takes input flattened from 2D patches to 1D time series and predicts long time-series sequences in a single forward operation, significantly enhancing inference speed. The process can be summarized as the following equations:
$$
\widehat{\boldsymbol{Y}}^{(i)}=W_5 \text { flatten }\left(\boldsymbol{O}_{\text {intra }}^{(i)}+\boldsymbol{O}_{\text {inter }}^{(i)}{ }^T+\boldsymbol{X}^{(i)}\right)
$$

\textbf{Weighted loss function:} Considering the varying importance of different variables in battery RUL prediction. To further enhance prediction performance, different from the loss function of traditional multivariate time series prediction, we propose to use the weighted mean square error as our loss function, which can be rewritten as:
$$
\text { WMSELoss }=\sum_{i=1}^C \alpha_i\left(\frac{1}{N} \sum_{j=1}^N\left(\widehat{\boldsymbol{Y}}_j^{(i)}-\boldsymbol{Y}_j^{(i)}\right)^2\right)
$$
where, $\alpha_i$ represents the weight assigned to the $i$-th variable, which is derived from the random forest regressor. It signifies the importance of each variable in predicting the Remaining Useful Life of the battery.

\subsection{The Multivariate Time Series}

In predicting the remaining useful life of rechargeable lithium-ion batteries, a lot of prior research uses only the capacity feature and barely considers more features of the battery in the charge-discharge cycles \cite{wang2022data} \cite{chen2022transformer}. However, the RUL of batteries can vary significantly from the same type of battery among different charge-discharge settings due to different temperatures, voltages, and currents. So, if we predict the RUL of batteries only relying on univariate time series, it will reduce the generalization ability of the prediction model.

To address these issues mentioned above, we propose a multivariate input representation that includes the features of charge and discharge cycles of rechargeable batteries. Scilicet, we predict the RUL of rechargeable batteries from a multivariate time series sequence. It is worth noting that, feature generation and selection are crucial to the prediction performance of our proposed approach. To better capture the evolving trends in time series data, we generate a feature by calculating the mean of the accumulated capacity for each discharge cycle. Moreover, the presence of noise and redundant information in the raw measurements can impede model convergence. To address this, we introduce a random forest regressor \cite{hwang2023feature} to identify and incorporate the most important features, thereby improving both the convergence speed and accuracy of the model. 

\subsection{Dual-Head MLP}

Motivated by PatchMixer \cite{gong2023patchmixer}, we employ the dual forecasting heads design in our IIP-Mixer model, including one intra-patch mixing MLP head and one inter-patch mixing MLP head. 

The intra-patch mixing MLP head comprises two fully connected layers and a GeLU nonlinearity, facilitating communication between time steps within a patch and capturing short-term temporal dependencies. The parameters within the intra-patch mixing MLP act as the short-term memory of IIP-Mixer, emphasizing the learning of information among local time steps without considering the entire input sequence. 

The structure of the inter-patch mixing MLP head just the same as the intra-patch mixing MLP head, allows communication between different patches and captures temporal dependencies across the whole input sequence. It consistently processes the same time step of each patch independently, extracting long-term temporal dependencies within the entire input sequence.

It is essential to note that, unlike PatchMixer, where serial dual heads are employed, our approach features parallel dual forecasting heads. The forecasting procedure simultaneously incorporates output information from the dual-head MLP and residual connections from past sequences to model future sequences using a linear projection. The training procedure of IIP-Mixer is summarized in Algorithm 1.

\begin{table}[!ht]
    \centering
    \begin{tabular}{ll} 
        \hline 
        \multicolumn{2}{l}{Algorithm 1 IIP-Mixer: PyTorch-like Pseudocode}\\
        \hline
        \multicolumn{2}{l}{$\#\ \mathrm{x}$ : segments of the univariate time series}
        \\        
        \multicolumn{2}{l}{$\#\ \mathrm{x}^{\top}:$ the transposition of input x }
        \\
        \multicolumn{2}{l}{$\#\ $params: parameters of the network: mixer$_{1:N}$ + fc}
        \\
        for x in loader: & \# load a minibatch
        \\
        \hspace{2em} $\mathrm{x}=\operatorname{tf}(\mathrm{x})$ & \# transform $\mathrm{x}$ from 1D to 2D  
        \\
        \hspace{2em} for $i$ in range(N) & \# loop number of mixer layer 
        \\
        \hspace{2em}\hspace{2em} x= mixer $_i(\mathrm{x})$ & \# the $i$-th mixer layer 
        \\
        \hspace{2em} $\mathrm{x}=$ ft(x) & \# flatten $\mathrm{x}$ from 2D to 1D 
        \\ 
        \hspace{2em} pred $=\mathrm{fc}(\mathrm{x})$ & \# full connection layer 
        \\
        \hspace{2em} loss $=$ WMSELoss $($ pred, true $)$ & \# weighted MSE loss function
        \\
        \hspace{2em} loss.backward() & \# back-propagate 
        \\
        \hspace{2em} update(params) & \# SGD update 
        \\ 
        \# mixer layers &  
        \\ 
        def mixer (x):&  
        \\ 
        \hspace{2em} $\mathrm{O}_{\text {intra }}=\mathrm{mlp}_{\text {intra }}(\mathrm{x})$ & \# intra-patch mixing MLP 
        \\ 
        \hspace{2em} $\mathrm{O}_{\text {inter }}=\mathrm{mlp}_{\text {inter }}(\mathrm{x^{\top}}) $& \# inter-patch mixing MLP 
        \\
        \hspace{2em} $ \mathrm{x}= \mathrm{O}_{\text {intra }}+\mathrm{O}_{\text {inter }}^{\top}+\mathrm{x}$ & \# aggregate output information
        \\
         \hspace{2em} return x & \\
        \hline 
    \end{tabular}
\end{table}

\section{Experiments}
\subsection{Datasets}
We conduct our experiments using PyTorch on the public dataset: NASA PCoE battery dataset, available from the NASA Ames Research Center website, contains the charge-discharge records of different lithium-ion batteries in multiple settings. Nonetheless, directly inputting battery measurements into a deep neural network poses challenges due to the potentially enormous amount of sampled data per cycle. In the realm of time series analysis, it is common practice to consolidate high-dimensional raw measurements at the cycle level through the application of statistical metrics, like minimum value, maximum value, and mean value at each cycle. 

\subsection{Baselines}
We benchmark our IIP-Mixer models against commonly used basic networks, including MLP, Transformer models, and its newly proposed variants such as DLinear \cite{zeng2023transformers} and Informer \cite{zhou2021informer}.
\begin{itemize}     
    \item MLP: A multilayer perceptron is just like a mathematical function that maps input values to output values. multiple layers are used to learn the dynamic and nonlinear degradation trend of the battery.
    \item Transformer: Transformer is a model that uses an attention mechanism for model training, it mainly consists of two components: an Encoder and a Decoder, with which we can predict the capacity degradation trend of the battery.
    \item Informer: A variant of transformer architecture, that efficiently handles extremely long input sequences by highlights dominating attention by halving cascading layer input. It predicts the long time-series sequences in one forward operation rather than a step-by-step way, which drastically improves the inference speed.
    \item DLinear: In consideration of permutation-invariant and “anti-ordering” to some extent of transformer-based architecture \cite{zeng2023transformers}, DLinear decomposes the time series into a trend and a remainder series and employs two one-layer linear networks to extract the temporal relations among an ordering set of continuous points.
\end{itemize}

\subsection{Implementation}
\subsubsection{Parameter Settings}
Our model has six key parameters: patch size, learning rate, dropout, length of patch, number of mixer blocks, and number of principal features of time series. During our experiments, the hyper-parameters are tuned based on the validation set. its details for the training process are shown in Table \ref{hps}. Additionally, for a more meaningful comparison, we aligned the size of lookback windows to 16, consistent with the approach in \cite{chen2022transformer}. 

\begin{table}[!ht]
\centering
\caption{Hyper-parameter summary.}
\label{hps}
\begin{tabular}{c|c} 
\hline
Hyper-Parameter & Range of Values  \\ 
\hline
Patch Size & \{2, 4, 8\} \\
Learning Rate &	\{0.0001, 0.0005, 0.001\} \\
Dropout & \{0.05, 0.1, 0.2\}\\
Length of Patch & \{2, 4, 8\} \\
\# of Mixer Blocks &	\{1, 2, 3, 4\} \\
\# of Principal Features & \{1,2,4,6,8,10,12,14,16\}\\
\hline
\end{tabular}
\end{table}

\subsubsection{Evaluation Metrics}
In our experiments, four evaluation metrics are used. In addition to three comment metrics Mean Absolute Error (MAE), Root Mean Square Error (RMSE), and Mean Absolute Percentage Error(MAPE), \cite{zhang2022method} we chose the Absolute Relative Error (ARE) \cite{chen2022transformer} to evaluate the prediction performance of battery RUL, which are defined as follow: 
$$
ARE = \frac{\left|RUL^{pre}-RUL^{ture}\right|}{RUL^{ture}} \times 100\%
$$
where $RUL^{ture}$ denotes the true RUL of the battery and $RUL^{pre}$ denotes the prediction of RUL from models. Additionally, we conducted each experiment three times using three consecutive seeds and reported the mean of the evaluation metrics.

\section{Results}
\subsection{Performance of IIP-Mixer}
Despite its simplicity, IIP-Mixer attains competitive results. To verify the performance of our methods, we conducted our experiments on four baseline methods. The MAE, RMSE, MAPE, and ARE scores obtained for all methods are shown in Table \ref{pfm}, the best results are shown in bold.
From the results we conclude the following: 
\begin{itemize}
\item The Transformer and its variant, the Informer model, excel in modeling both long-term and short-term dependencies, showcasing superior performance on long-time series data. However, it is worth noting that, they can easily lead to overfitting for small datasets, such as the NASA PCoE battery dataset. 
\item The MPL model is adept at capturing global temporal patterns but may struggle to capture local temporal patterns from time series data. Consequently, its performance remains average across all evaluation metrics compared to other methods.
\item The DLinear model decomposes the time series into a trend and a remainder series. It utilizes two one-layer linear networks to extract temporal relations among an ordered set of continuous points, making it adept at capturing both the trend and season of a time series.
\item  Among all the baseline methods, our proposed model IIP-Mixer achieves the best experimental results. It demonstrates that IIP-Mixer can capture the local and global temporal patterns in time series data, this is a great help for battery RUL prediction. 
\end{itemize}

\begin{table}[!ht]
\centering
\caption{Performances of methods on NASA PCoE battery dataset.}
\label{pfm}
\begin{tabular}{ccccc} 
\hline
Methods & MAE(Ah) &	RMSE(Ah) & MAPE(\%) & ARE(\%)  \\ 
\hline
Transformer&0.055&0.073&3.697&9.589  \\
Informer&0.049&0.063&3.281&4.110\\
MLP&0.050&0.066&3.402&6.393\\
DLinear&0.041&0.052&2.732&3.196\\
IIP-Mixer& \textbf{0.037}&\textbf{0.048}&\textbf{2.480}&\textbf{1.370}\\
\hline
\end{tabular}
\end{table}

\subsection{Computation Efficiency}
We investigated the computational efficiency of all neural networks mentioned above based on the length $L$ of the input time series. Comparisons of theoretical time complexity and memory usage \cite{zhou2021informer} are summarized in Table \ref{cps}. It is important to note that, in our model, the patch size $W$ can equal the patch number $H$, so we have $W\times H = W^2 = H^2 = L$. It is evident that our model, like DLinear, has the lowest cost across all computational metrics.

\begin{table}[!ht]
\centering
\caption{$L$-related computation statics of each layer.}
\label{cps}
\begin{tabular}{cccc} 
\hline
\multirow{2}{4em}{Methods}& \multicolumn{2}{c}{Training}&Testing  \\
&Time&Memory&Step
\\
\hline
 Transformer& $\mathcal{O}(L^2)$ & $\mathcal{O}(L^2)$ & $L$
\\
 Informer& $\mathcal{O}(L\log L)$ & $\mathcal{O}(L\log L)$ & $1$
\\
 MLP& $\mathcal{O}(L^2)$ & $\mathcal{O}(L^2)$ & $1$
\\
 DLinear& $\mathcal{O}(L)$ & $\mathcal{O}(L)$ & $1$
\\
 IIP-Mixer& $\mathcal{O}(L)$ & $\mathcal{O}(L)$ & $1$
\\
\hline
\end{tabular}
\end{table}

\subsection{Ablation Study}
\subsubsection{Effect of Dual-Head MLP}
To evaluate the effectiveness of dual-head MLP, we individually remove the intra-patch mixing MLP head or inter-patch mixing MLP head. Table \ref{dhmlp} demonstrates that the dual-head MLP mechanism outperforms all other configurations. This result highlights the effectiveness of the dual-head mechanism in comparison to a single output head. 

It is worth noting that, the performance of architecture without an intra-patch mixing MLP head is better than that without inter-patch mixing MLP head. Scilicet, inter-patch mixing MLP that captures global temporal patterns performs better than intra-patch mixing MLP that captures local temporal patterns in battery RUL prediction.

\begin{table}[!ht]
\centering
\caption{Ablation study of dual-head MLP.}
\label{dhmlp}
\begin{tabular}{ccccc} 
\hline
Methods & MAE(Ah) &	RMSE(Ah) & MAPE(\%) & ARE(\%)  \\ 
\hline
w/o inter&0.081&0.095&5.443&34.703 \\
w/o intra&0.044&0.055&2.940&5.023 \\
IIP-Mixer&\textbf{0.037}&\textbf{0.048}&\textbf{2.480}&\textbf{1.370} \\
\hline
\end{tabular}
\end{table}

\subsubsection{Serial vs Parallel Heads}
Unlike the serial structure of dual-head MLP in MLP-Mixer and PatchMixer, we introduce a parallel structure for dual heads. To evaluate the effectiveness of the parallel heads structure, we compare the performance of serial heads (intra-first), serial heads (inter-first), and parallel heads. As shown in Table \ref{sph}, it is evident that the parallel heads structure outperforms the serial structure in battery RUL prediction.

\begin{table}[!ht]
\centering
\caption{Comparison of serial and parallel heads.}
\label{sph}
\begin{tabular}{ccccc} 
\hline
Methods & MAE(Ah) &	RMSE(Ah) & MAPE(\%) & ARE(\%)  \\ 
\hline
Serial Heads
(inter-first)&0.063&0.076&4.258&15.068
 \\
Serial Heads
(intra-first)&0.045&0.056&3.032&4.110
\\
Parallel Heads&\textbf{0.037}&\textbf{0.048}&\textbf{2.480}&\textbf{1.370} \\
\hline
\end{tabular}
\end{table}

\subsubsection{Effect of Weighted Loss Function}
In the majority of recent research on MLP-based models, the loss functions have not taken into account the varying importance of different variables. As shown in Table \ref{ww}, it is evident that the weighted loss function outperforms the loss function without weighting, resulting in relative improvements of 5\%, 2\%, and 3\% in Mean Absolute Error (MAE), Root Mean Square Error (RMSE) and Mean Absolute Percentage Error (MAPE), respectively.

\begin{table}[!ht]
\centering
\caption{Comparison of loss function: with weighted vs. w/o weighted.}
\label{ww}
\begin{tabular}{ccccc} 
\hline
Methods & MAE(Ah) &	RMSE(Ah) & MAPE(\%) & ARE(\%)  \\ 
\hline
w/o weighted&0.039&0.049&2.562&1.370 
 \\
w weighted&\textbf{0.037}&\textbf{0.048}&\textbf{2.480}&\textbf{1.370} \\
\hline
\end{tabular}
\end{table}

\subsubsection{Effect of Multivariate}
We investigate the impact of multivariate time series in Table \ref{uvsm}. The learning performance with principal multivariate time series surpasses that of univariate time series. Specifically, predicting the Remaining Useful Life of batteries using a multivariate time series of principal features significantly enhances the generalization ability of the prediction model. 

It's important to highlight that training with multivariate time series containing all features may lead to a reduction in model performance. Therefore, the feature selection pipeline plays a critical role in determining prediction performance.

\begin{table}[!ht]
\centering
\caption{Comparison of time series: univariate vs. multivariate.}
\label{uvsm}
\begin{tabular}{ccccc} 
\hline
Methods & MAE(Ah) &	RMSE(Ah) & MAPE(\%) & ARE(\%)  \\ 
\hline
Univariate&0.042&0.054&2.822&3.653
 \\
Multivariate(full)&0.052&0.064&3.510&7.306
\\
Multivariate
(principal)
&\textbf{0.037}&\textbf{0.048}&\textbf{2.480}&\textbf{1.370} \\
\hline
\end{tabular}
\end{table}

\section{Conclusion}
In this paper, we present IIP-Mixer, an innovative MLP-Mixer-based architecture designed to predict Remaining Useful Life in batteries. IIP-Mixer incorporates parallel dual-head MLP: the intra-patch mixing MLP and inter-patch mixing MLP. The intra-patch mixing MLP independently applies MLP to each patch, capturing local temporal patterns in the short-term period. On the other hand, the inter-patch mixing MLP applies MLP across all patches from the input sequence, capturing global temporal patterns in the long-term period. Moreover, recognizing the varying importance of features in RUL prediction, we propose a weighted mean square error loss function to enhance prediction accuracy. Our experiments demonstrate that IIP-Mixer achieves competitive performance in battery RUL prediction, outperforming other popular time-series frameworks.

\section*{Acknowledgment}
This research was funded in part by the Science and Technology Development Fund, Macau SAR, grant number 0093/2022/A2, 0076/2022/A2, and 0008/2022/AGJ. It was also supported by the Guangdong Provincial Department of Education's Key Special Projects, with project numbers 2022ZDZX1073 and 2023ZDZX1086, as well as by the Special Fund for Dongguan’s Rural Revitalization Strategy, under number 20211800400102. Furthermore, support was provided by the Dongguan Sci-tech Commissioner Program, with grant numbers 20221800500842, 20221800500632, 20221800500822, 20221800500792, and 20231800500442. Additionally, it received support from the Dongguan Science and Technology of Social Development Program, under number 20231800936942. Finally, funding was provided by the Dongguan Songshan Lake Enterprise Special Envoy Project.

\printbibliography

@article{wei2021remaining,
  title={Remaining useful life prediction of lithium-ion batteries based on Monte Carlo Dropout and gated recurrent unit},
  author={Wei, Meng and Gu, Hairong and Ye, Min and Wang, Qiao and Xu, Xinxin and Wu, Chenguang},
  journal={Energy Reports},
  volume={7},
  pages={2862--2871},
  year={2021},
  publisher={Elsevier}
}

@article{alsuwian2024review,
  title={A Review of Expert Hybrid and Co-Estimation Techniques for SOH and RUL Estimation in Battery Management System with Electric Vehicle Application},
  author={Alsuwian, Turki and Ansari, Shaheer and Zainuri, Muhammad Ammirrul Atiqi Mohd and Ayob, Afida and Hussain, Aini and Lipu, MS Hossain and Alhawari, Adam RH and Almawgani, AHM and Almasabi, Saleh and Hindi, Ayman Taher},
  journal={Expert Systems with Applications},
  pages={123123},
  year={2024},
  publisher={Elsevier}
}

@article{xu2023hybrid,
  title={A hybrid ensemble deep learning approach for early prediction of battery remaining useful life},
  author={Xu, Qing and Wu, Min and Khoo, Edwin and Chen, Zhenghua and Li, Xiaoli},
  journal={IEEE/CAA Journal of Automatica Sinica},
  volume={10},
  number={1},
  pages={177--187},
  year={2023},
  publisher={IEEE}
}

@article{santhanagopalan2006review,
  title={Review of models for predicting the cycling performance of lithium-ion batteries},
  author={Santhanagopalan, Shriram and Guo, Qingzhi and Ramadass, Premanand and White, Ralph E},
  journal={Journal of power sources},
  volume={156},
  number={2},
  pages={620--628},
  year={2006},
  publisher={Elsevier}
}

@article{kemper2015simplification,
  title={Simplification of pseudo two-dimensional battery model using dynamic profile of lithium concentration},
  author={Kemper, Paulo and Li, Shengbo Eben and Kum, Dongsuk},
  journal={Journal of Power Sources},
  volume={286},
  pages={510--525},
  year={2015},
  publisher={Elsevier}
}

@article{wang2022data,
  title={A data-driven method with mode decomposition mechanism for remaining useful life prediction of lithium-ion batteries},
  author={Wang, Jianguo and Zhang, Shude and Li, Chenyu and Wu, Lifeng and Wang, Yingzhou},
  journal={IEEE Transactions on Power Electronics},
  volume={37},
  number={11},
  pages={13684--13695},
  year={2022},
  publisher={IEEE}
}

@article{wang2023improved,
  title={Improved anti-noise adaptive long short-term memory neural network modeling for the robust remaining useful life prediction of lithium-ion batteries},
  author={Wang, Shunli and Fan, Yongcun and Jin, Siyu and Takyi-Aninakwa, Paul and Fernandez, Carlos},
  journal={Reliability Engineering \& System Safety},
  volume={230},
  pages={108920},
  year={2023},
  publisher={Elsevier}
}

@article{zhang2023data,
  title={A data-model interactive remaining useful life prediction approach of lithium-ion batteries based on PF-BiGRU-TSAM},
  author={Zhang, Jiusi and Huang, Congsheng and Chow, Mo-Yuen and Li, Xiang and Tian, Jilun and Luo, Hao and Yin, Shen},
  journal={IEEE Transactions on Industrial Informatics},
  year={2023},
  publisher={IEEE}
}

@article{chen2022transformer,
  title={Transformer network for remaining useful life prediction of lithium-ion batteries},
  author={Chen, Daoquan and Hong, Weicong and Zhou, Xiuze},
  journal={Ieee Access},
  volume={10},
  pages={19621--19628},
  year={2022},
  publisher={IEEE}
}

@inproceedings{ye2023prediction,
  title={Prediction with time-series mixer for the S\&P500 index},
  author={Ye, Junyi and Gu, Jingyi and Dash, Ankan and Deek, Fadi P and Wang, Guiling Grace},
  booktitle={2023 IEEE 39th International Conference on Data Engineering Workshops (ICDEW)},
  pages={20--27},
  year={2023},
  organization={IEEE}
}

@inproceedings{zeng2023transformers,
  title={Are transformers effective for time series forecasting?},
  author={Zeng, Ailing and Chen, Muxi and Zhang, Lei and Xu, Qiang},
  booktitle={Proceedings of the AAAI conference on artificial intelligence},
  volume={37},
  number={9},
  pages={11121--11128},
  year={2023}
}

@article{wei2024state,
  title={State of health and remaining useful life prediction of lithium-ion batteries with conditional graph convolutional network},
  author={Wei, Yupeng and Wu, Dazhong},
  journal={Expert Systems with Applications},
  volume={238},
  pages={122041},
  year={2024},
  publisher={Elsevier}
}

@article{tolstikhin2021mlp,
  title={Mlp-mixer: An all-mlp architecture for vision},
  author={Tolstikhin, Ilya O and Houlsby, Neil and Kolesnikov, Alexander and Beyer, Lucas and Zhai, Xiaohua and Unterthiner, Thomas and Yung, Jessica and Steiner, Andreas and Keysers, Daniel and Uszkoreit, Jakob and others},
  journal={Advances in neural information processing systems},
  volume={34},
  pages={24261--24272},
  year={2021}
}

@article{liang2024hybrid,
  title={A hybrid approach based on neural network and double exponential model for remaining useful life prediction},
  author={Liang, Junyuan and Liu, Hui and Xiao, Ning-Cong},
  journal={Expert Systems with Applications},
  pages={123563},
  year={2024},
  publisher={Elsevier}
}

@article{wu2016online,
  title={An online method for lithium-ion battery remaining useful life estimation using importance sampling and neural networks},
  author={Wu, Ji and Zhang, Chenbin and Chen, Zonghai},
  journal={Applied energy},
  volume={173},
  pages={134--140},
  year={2016},
  publisher={Elsevier}
}

@article{sutskever2014sequence,
  title={Sequence to sequence learning with neural networks},
  author={Sutskever, Ilya and Vinyals, Oriol and Le, Quoc V},
  journal={Advances in neural information processing systems},
  volume={27},
  year={2014}
}

@article{zhang2018long,
  title={Long short-term memory recurrent neural network for remaining useful life prediction of lithium-ion batteries},
  author={Zhang, Yongzhi and Xiong, Rui and He, Hongwen and Pecht, Michael G},
  journal={IEEE Transactions on Vehicular Technology},
  volume={67},
  number={7},
  pages={5695--5705},
  year={2018},
  publisher={IEEE}
}

@article{shi2021dual,
  title={A dual-LSTM framework combining change point detection and remaining useful life prediction},
  author={Shi, Zunya and Chehade, Abdallah},
  journal={Reliability Engineering \& System Safety},
  volume={205},
  pages={107257},
  year={2021},
  publisher={Elsevier}
}

@article{vaswani2017attention,
  title={Attention is all you need},
  author={Vaswani, Ashish and Shazeer, Noam and Parmar, Niki and Uszkoreit, Jakob and Jones, Llion and Gomez, Aidan N and Kaiser, {\L}ukasz and Polosukhin, Illia},
  journal={Advances in neural information processing systems},
  volume={30},
  year={2017}
}

@article{chen2023tsmixer,
  title={Tsmixer: An all-mlp architecture for time series forecasting},
  author={Chen, Si-An and Li, Chun-Liang and Yoder, Nate and Arik, Sercan O and Pfister, Tomas},
  journal={arXiv preprint arXiv:2303.06053},
  year={2023}
}

@article{hwang2023feature,
  title={Feature importance measures from random forest regressor using near-infrared spectra for predicting carbonization characteristics of kraft lignin-derived hydrochar},
  author={Hwang, Sung-Wook and Chung, Hyunwoo and Lee, Taekyeong and Kim, Jungkyu and Kim, YunJin and Kim, Jong-Chan and Kwak, Hyo Won and Choi, In-Gyu and Yeo, Hwanmyeong},
  journal={Journal of Wood Science},
  volume={69},
  number={1},
  pages={1},
  year={2023},
  publisher={Springer}
}

@article{gong2023patchmixer,
  title={Patchmixer: A patch-mixing architecture for long-term time series forecasting},
  author={Gong, Zeying and Tang, Yujin and Liang, Junwei},
  journal={arXiv preprint arXiv: 2310.00655},
  year={2023}
}

@inproceedings{zhou2021informer,
  title={Informer: Beyond efficient transformer for long sequence time-series forecasting},
  author={Zhou, Haoyi and Zhang, Shanghang and Peng, Jieqi and Zhang, Shuai and Li, Jianxin and Xiong, Hui and Zhang, Wancai},
  booktitle={Proceedings of the AAAI conference on artificial intelligence},
  volume={35},
  number={12},
  pages={11106--11115},
  year={2021}
}

@article{zhang2022method,
  title={A method for capacity prediction of lithium-ion batteries under small sample conditions},
  author={Zhang, Meng and Kang, Guoqing and Wu, Lifeng and Guan, Yong},
  journal={Energy},
  volume={238},
  pages={122094},
  year={2022},
  publisher={Elsevier}
}
\end{document}